\documentclass{article}

\usepackage{microtype}
\usepackage{graphicx}
\usepackage{subcaption}
\usepackage{booktabs} 
\usepackage{hyperref}

\usepackage[preprint]{icml2026}

\usepackage{amsmath}
\usepackage{amssymb}
\usepackage{mathtools}
\usepackage{amsthm}
\usepackage{tabularx}

\usepackage[capitalize,noabbrev]{cleveref}

\theoremstyle{plain}

\theoremstyle{definition}

\theoremstyle{remark}

\usepackage[textsize=tiny]{todonotes}
\usepackage{tikz}
\usetikzlibrary{arrows.meta,positioning,calc}

\icmltitlerunning{}

\begin{document}

\twocolumn[
  \icmltitle{\texorpdfstring{Self-Review Reinforcement Learning (SRRL)\\with Cross-Episode Memory and Policy Distillation}{Self-Review Reinforcement Learning (SRRL) with Cross-Episode Memory and Policy Distillation}}




    \begin{icmlauthorlist}
      \icmlauthor{Muhammad Zain Amin}{ets,sc}
      \icmlauthor{Kibele Sebnem Yildirim}{mcgill}
    \end{icmlauthorlist}
    
    \icmlaffiliation{ets}{Department of Software Engineering and IT, École de Technologie Supérieure, Montreal, Canada}
    \icmlaffiliation{sc}{Centre intégré de traumatologie, Hôpital du Sacré-Cœur de Montréal, Université de Montréal, Montreal, Canada} 
    \icmlaffiliation{mcgill}{Desautels Faculty of Management, McGill University, Montreal, Canada}
    
    \icmlcorrespondingauthor{Muhammad Zain Amin}{muhammad-zain.amin.1@ens.etsmtl.ca}
    \icmlcorrespondingauthor{Kibele Sebnem Yildirim}{kibele.yildirim@mail.mcgill.ca}

  \icmlkeywords{Machine Learning, ICML}

  \vskip 0.3in

]



\begingroup\makeatletter\let\Hy@Warning\@gobble\printAffiliationsAndNotice{}\endgroup

\begin{abstract}

Reinforcement Learning is commonly used to train large language models using environmental feedback. In applied settings, the environment usually provides sparse or delayed feedback. This makes it difficult for the model to pinpoint which actions in its reasoning led to success or failure. So, learning effectively from these signals is hard because the model must determine how each failure should inform meaningful behavioral corrections in subsequent iterations. We introduce a training framework, \textit{Self-Review Reinforcement Learning}, that embeds an explicit \textit{self-review} step into each RL episode. When a first-pass response fails, the model generates a \textit{self-review} to identify what went wrong, which conditions an improved second attempt. Unlike inference-time reflection approaches, such as \textit{Reflexion}, the framework optimizes self-review with policy gradients and internalizes improvements into the base policy via \textit{selective distillation}, ensuring they persist across future episodes. A \textit{cross-episode memory} keeps successful self-reviews for reuse when encountering similar tasks in future episodes during training. We evaluate SRRL against a standard RLVR baseline using the GRPO optimizer across two language models, Qwen 3-4B and OLMo-3-7B, on GSM8K benchmark. SRRL consistently outperforms the RLVR in final reward performance and achieves greater learning efficiency by successfully transforming feedback into behavioral improvement.
Code github repository: \href{https://github.com/ZainAmin/COMP767-Project}{https://github.com/ZainAmin/COMP767-Project}

\end{abstract}

\section{Introduction}

\begin{center}
\captionsetup{hypcap=false}
\includegraphics[width=\columnwidth,height=0.9\textheight,keepaspectratio]{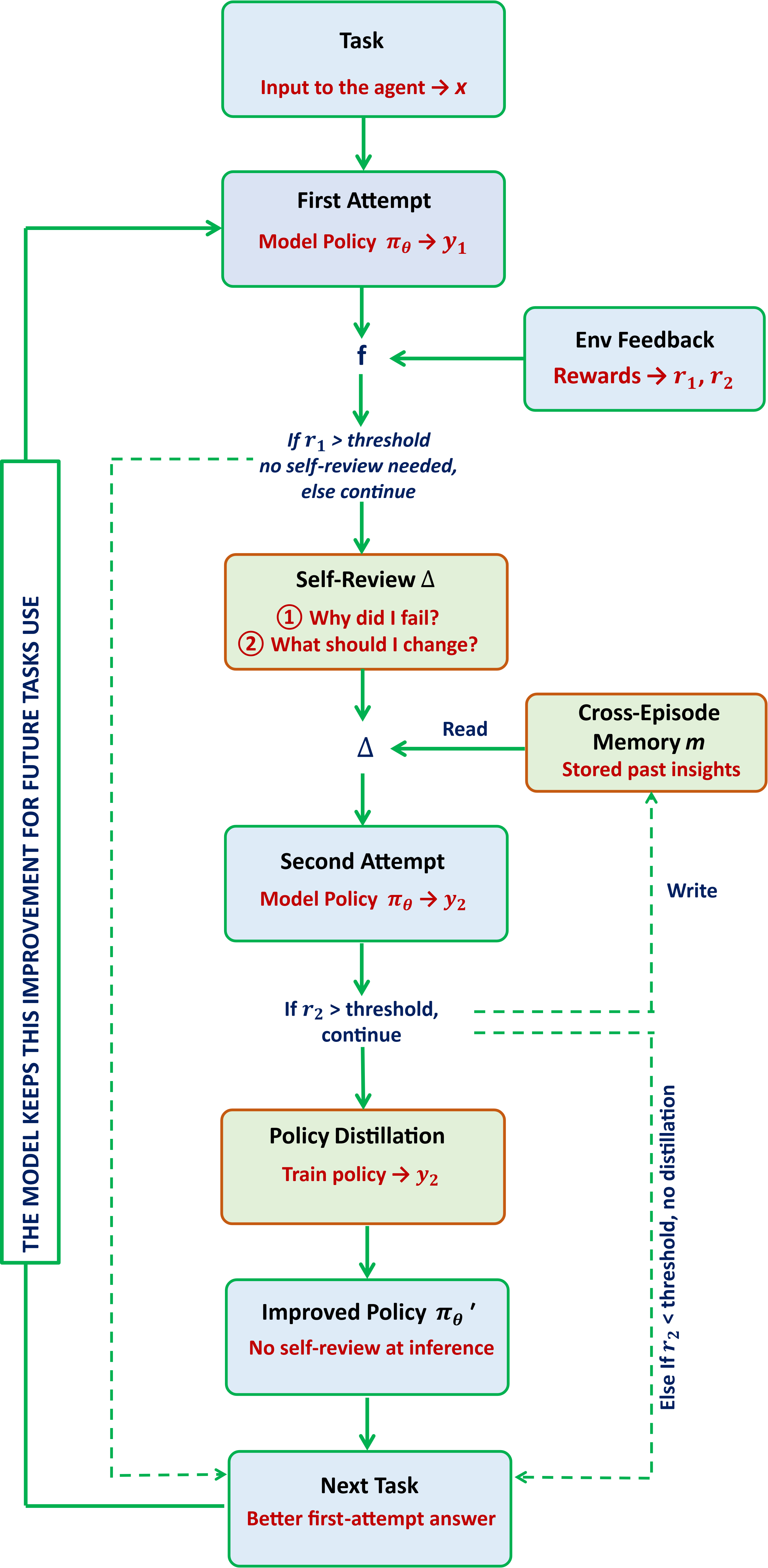}
\captionof{figure}{The SRRL training framework}
\label{fig:srrl_overview}
\end{center}

These days, large language models (LLMs) are increasingly employed as decision-making agents in situations with delayed rewards and partial knowledge \citep{bai2026kimi,song2026textfeedback,wang2025cogito,zhang2025earlyexperience}.

Reinforcement Learning provides an intuitive framework for enhancing such agents. After the model agent generates a first-pass response, it typically receives environmental feedback in the form of outcome rewards. Environmental feedback is the information or signal the model receives from the world outside itself in response to its actions. But the problem is that the environment often provides no feedback for individual steps, making the feedback rewards delayed and sparse. Therefore, the model is confused about how to choose the correct stepwise reasoning to fix its behaviour and ends up being trained inefficiently with respect to sampling  \citep{shi2025wildfeedback, zhang2025earlyexperience}. Especially in multistep reasoning, errors in intermediate steps can invalidate the entire solution trajectories and credit assignments, leaving the model unable to recover.

In supervised fine-tuning (SFT), the model’s base policy is trained by learning to reproduce human-provided fixed examples. During training, the model observes pairs of prompts and their desired responses, then adjusts its parameters to minimise the difference between the generated and desired outputs. This kind of imitation learning allows the model to become highly effective at generating highly realistic, well-structured responses it has seen before. However, this strength also creates a fundamental problem. SFT is applied to a static dataset, and the model lacks an inherent mechanism to revise its behavior. In real world use cases, SFT doesn’t have the ability to incorporate feedback and learn from its mistakes; instead it continues to rely on the training patterns it encoded earlier.

Reinforcement learning with verifiable rewards (RLVR) takes the model training into interactive contexts. RLVR allows models to progress through a trial-and-error approach while receiving external validation feedback. In the RLVR framework, agents take an action and receive a scalar reward for success, and over time, update their policy to increase the probability of high-reward outcomes. This process takes one step beyond imitation learning and allows models to adapt and refine their performance over time with experience. The key limitation is that the reward signals generally provide a numerical indication of success or failure without explaining why the action was effective or incorrect. As a result, the agents continue to make similar mistakes and adopt suboptimal strategies, without any stable corrections that persist across episodes. For this reason, the corrective structure of the good responses taken by the agent must be deduced directly from noisy, sparse, or delayed rewards.

A natural progression from here is to do structure learning around experience itself by transforming the feedback into intermediate reasoning representation, such as review summaries, error analysis and the strategy updates, which can help make the corrective insights straightforward. By integrating the intermediate learning representation, agents’ learning experience can reliably consolidate the improvements, as well as adjust their behavior more accurately without totally relying on reward-driven exploration.

\section{Literature review}

\subsection{RLs and LLMs:}

The concept of learning through experience has become widely central to the development of large language models. While early research shows that the models refine their outputs and align their behaviour based on human-provided scores and evaluations, they have evolved into Reinforcement Learning from Human Feedback (RLHF) \citep{ouyang2022instructgpt, christiano2017preferences, shi2024safer, shinn2023reflexion}.

Recent research has extended reinforcement learning beyond the idea of preference alignment in order to enhance the mathematical and symbolic reasoning of the LLMs. Instead of just relying on human feedback, the models receive the rewards directly from the programmatically verifiable signals. For example, a code snippet can be executed against the test cases, a mathematical answer can be checked against proven ground truth, etc. These checks actually provide the structured supervision signals that evaluate the reasoning outcomes. The key shift in these strategies is the replacement of subjective reward models with deterministic mechanism processes  \citep{guo2025deepseekr1, song2026textfeedback,openai2024o1}.

Similarly, the agent-oriented research on the LLMs presents the models as an interactive policy in order to operate in an external environment instead of just generating a single static response. In this approach, models continue to take actions, gather observations and task-dependent rewards in order to solve problems with multitask reasoning \citep{yao2023react, jin2025searchr1, bai2026kimi}.

Although these approaches differ, ranging from preference alignment to formal reasoning and interactive problem solving, all of them share the same training paradigm, which is that the environmental feedback is reduced to a scalar reward, which is optimised through gradient policy updates. This limitation requires the model to implicitly infer how to correct its mistakes through repeated exploration rather then explicitly guided towards the structured revision of its reasoning process. As a result of this limitation, our SRRL framework departs from the reward-driven optimisation towards the process where the feedback is transformed into structured analysis and behaviour revision before being internalised into the base policy.

\subsection{Learning via Trails:}

A large number of research studies suggest that the next scaling era for AI will be less dependent on statistical human feedback and more on the interaction of agents, which streamlines the learning via trials. This way, agents can easily produce experience-rich data through continual interaction with their environment. One of the leading studies by Silver and Sutton’s lab argues that data streams generated by agents’ interaction, combined with long-term decision making, can help the models to go beyond imitation and adopt learning via trails mechanism \citep{silver2025experience}. This study gives an inspiration to adopt strategies that can convert the failed trials into usable learning signals instead of just relying on the successful trials.

Earlier work on simple reinforcement learning handles the challenge of sparse rewards. The issue was that the agents only received feedback after successful trials, and it was solved by introducing a mechanism that also extracts a signal from failed trials. A notable contribution is the goal relabeling, as proposed in Hindsight Experience Replay (HER) \citep{andrychowicz2018her}. In the sparse-rewards setting, an agent can attempt to succeed, but if it fails, it receives no rewards, providing very little to no guidance for improvement. HER solves this issue by relabeling the goal of a failed trial as the outcome that was actually achieved. It means that even if the agent did not achieve the intended goal, the trial is still relabeled as a success for a different achieved state. So this whole process ultimately improves sample efficiency by converting failed trials into useful training data.

Similarly, in the LLM-agent setting environment, the author addresses the void between the reward-driven reinforcement learning and the imitation learning by allowing the agents to train from their own interaction traces, even when the rewards are sparse or not available. The agent, instead of relying on the external feedback, uses its generated future states and decisions as a form of self-supervision. Self-reflection also plays an important role as a way to observe suboptimal or failed behaviour \citep{zhang2025earlyexperience}.

The inference-time reflection methods demonstrate that LLMs can improve their responses by revising them before consolidating and producing a final response. The authors come to the agreement that all the self-revision techniques which give the model the ability to think twice, help the model learn from its own mistakes and correct its behaviour, often lead to improved performance. But these approaches require the ’think twice’ step to be executed during deployment. For this reason, the model must access memory during inference, which increases computational cost and restricts improvements in runtime framing rather than permanent changes in the policy  \citep{zelikman2022star, madaan2023selfrefine, shinn2023reflexion}.

In contrast, concurrent research aims to incorporate conditioned feedback directly into the training process rather than relying on inference-time self-revision. Recently, \citet{hubotter2026selfdistill} \citet{song2026textfeedback} used reinforcement learning with textual feedback. The feedback-conditioned teacher policy is distilled into the student policy. It doesn’t require direct feedback during deployment. Through distillation, the student learns to mimic the teacher’s feedback-informed decisions, thereby internalising the corrective direction within its parameters.

Our proposed SRRL also aligns with this line of work. It places focus on the self-review during training. After the first-pass response fails, the model generates a structured self-review that guides a refined retry in the same episode. The proposed approach treats the self-review as a structured credit assignment task. By combining self-review with cross-episode memory and policy distillation, the proposed criteria transform the experience into a lasting policy improvement without requiring self-review at inference time.

\section{Self-Review Reinforcement Learning}

Self-Review Reinforcement Learning is a training framework which is designed to use the environmental feedback for structured behavioral corrections. In standard RLVR, agents usually receive a scalar reward after generating a response, and right away update their policy parameters from this signal afterwards. The main limitation of RLVR is that the reward provides no mechanism for the agent to know which intermediate reasoning steps were responsible for the failure or the success. This problem forces the model to rediscover corrective strategies through repeated exploration, which further causes the inefficient learning problem. SRRL addresses these problems by proposing an explicit self-review step within each training episode. This process is anchored by two supporting mechanisms, cross-episode memory and policy distillation, which ensure that the insights produced through self-review accumulate and persist across training.

Each SRRL training episode proceeds through several stages, as shown in Figure~\ref{fig:srrl_overview}. Given an input problem $x$, the model $\pi_\theta$ generates the first-pass response $y_1$. Afterwards, the environment evaluates the $y_1$ response and returns a scalar outcome reward $r_1 \in \{0,1\}$ along with textual feedback $f_1$. If the $r_1$ falls below a reward threshold $\tau$ (i.e., $r_1 < \tau$), then the model enters a self-review phase, an explicit analysis of what went wrong in response $y_1$ and what corrective strategies can be applied to correct or improve the response. Conditioned on $x$, $y_1$, $r_1$, $f_1$, and any retrieved memory $m$ from the cross-episode memory mechanism, the model generates a second-pass response $y_2$. The environment evaluates $y_2$ and returns $r_2$, $f_2$. If the response $y_2$ succeeds (i.e., $r_2 > \tau$), the self-review is written in the cross-episode memory $m$, and the episode ends with a distillation update that trains the base policy to directly produce $y_2$ quality responses from $x$ alone. This eliminates the redundant exploration problem of corrective strategies for similar tasks in subsequent training episodes.

The design of SRRL is beneficial as a key deployment property, as stated below:

\[
y \sim \pi_{\theta}(\cdot \mid x)
\]

meaning that at inference time, the model does not require any self-review, memory retrieval, or policy distillation. All improvements accumulated during training through SRRL are encoded into the base policy weights, which effectively ensure zero latency at deployment. This factor directly addresses the limitation of inference-time self-reflection methods, such as Reflexion \citep{shinn2023reflexion} and Self-Refine \citep{madaan2023selfrefine}, which require reflection to be present during deployment.

In SRRL, the self-review is not triggered automatically. It is activated only if the first-pass response falls below the reward threshold. This condition is very important for two reasons, first is that by applying self-review to already correct response trajectories cause the reward hacking problem, which precisely means that the model can easily generate instance-specific shortcuts which can guarantee the success on current sample but will be unable to generalize. Secondly, early in training, when the first-pass solution rewards are usually low, the self-review heavy batches become dominated by off-policy second-pass attempt data that ultimately weakens the on-policy learning signal and destablizes optimization.

By restricting the self-review mechanism to failed trajectories, we ensure that successful responses remain purely on-policy and each training batch retains a stable on-policy component. This factor directly aligns with our model-deployment property, since the model has no access to self-review during deployment; training without such conditioning would widen the gap between training-time and deployment-time behavior. When the condition occurs, the model generates the self-review as

\[
\Delta \sim \pi_\theta(x, y_1, r_1, f_1, m)
\]

After self-review $\Delta$, the model $\pi_\theta$ attempts the second-pass solution as

\[
y_2 \sim \pi_\theta(x, \Delta)
\]

Moreover, a single self-review only provides a learning signal within the training episode in which it is generated. Therefore, without an additional mechanism, it is impossible to retain these insights and the model has to rediscover them independently each time for structurally similar tasks in future episodes. To overcome this limitation, we incorporate a cross-episode memory step and use it as retrieval context for future training episodes.

The cross-episode memory stores successful self-reviews as a plain text record. The memory item $m$ is directly provided to the model $\pi_\theta$ as part of the system context during the self-review $\Delta$ step. The successful self-review $\Delta$ is stored in $m$ only if the second-pass attempt succeeds, as follows:

\[
m_{t+1} =
\begin{cases}
m_t \cup \{\Delta\}, & \text{if } r_2 > \tau,\\
m_t, & \text{otherwise.}
\end{cases}
\]

This selective $\Delta$ storing ensures that only the validated corrective strategies persist. Subsequently, the selective supervised policy distillation step is applied by training the base policy on the successful second-attempt response. We optimize the model to directly reproduce $y_2$-level responses (i.e., reproduce improved behavior directly) from $x$ without access to self-review $\Delta$ context during future training episodes for structurally similar problems and at deployment as well.

\[
L_{\mathrm{distill}}(\theta) = - \mathbb{E}_{(x, y_2, r_2) \sim \mathcal{D}} \left[ \mathbb{I}_{\{r_2 > \tau\}} \log \pi_\theta(y_2 \mid x) \right]
\]

This $I$ function triggers the distillation update only to successful second attempts. Failed second pass solution attempts do not contribute to the distillation loss.

\subsection{RL Objective}

Both the first pass, the second pass attempts, and self-review are optimized with the RL objective. We adopt GRPO (Group Relative Policy Optimization) \citep{deepseekmath} as the underlying gradient policy optimizer. The policy loss is defined as:

\[
L_{\mathrm{policy}}(\theta) = - \mathbb{E}_{(x,y) \sim \mathcal{D}} \left[ A(x, y) \log \pi_\theta(y \mid x) \right]
\]

Here, $y$ represents the model output, and the conditioning context corresponds to the model inputs for each generation step. The advantage estimate $A$ is calculated from associated outcome rewards using GRPO group-relative normalization; for a group of $G$ sampled responses, the advantages are normalized within the group.

\[
A_i = \frac{r_i - \mathrm{mean}_j\{r_j\}}{\mathrm{std}_j\{r_j\}}
\]

The self-review $\Delta$ is assigned the reward $\tilde{r} = r_2$; that means, the self-review is rewarded by the downstream success it produces.

In order to prevent the policy from drifting too far from the reference model under the combined gradient signals, we apply KL regularization with a low-variance KL loss and a fixed coefficient of $\beta = 0.001$.

\subsection{Combined Objective}

The full SRRL training objective per episode combines policy gradient loss over all generated outputs with the selective distillation loss:

\[
L_{\mathrm{SRRL}} = L_{\mathrm{policy}} + L_{\mathrm{distill}} + \beta \cdot L_{\mathrm{KL}}
\]

The $L_{\mathrm{policy}}$ term covers the first-pass attempt unconditionally, and the self-review and the second-pass attempt are activated only when this condition is triggered (i.e., $r_1 < \tau$). The $L_{\mathrm{distill}}$ term is masked to zero in episodes where $r_2 < \tau$. This means that failed episodes still contribute an RL signal via $L_{\mathrm{policy}}$, but do not contaminate the distillation process with unsuccessful strategies. Improvements are fully encoded in the model weights, preserving the gains of self-review-driven training at zero additional deployment cost.

\section{Experiments}

We evaluate SRRL against a standard RLVR baseline on GSM8K agentic reasoning benchmark.

\subsection{Task}

Task is chosen to provide meaningful reward signal for the base model. We also ensure that the models have partial competence so that GRPO can learn from variance in rewards. We use GSM8K, a grade-school mathematics word-problem reasoning benchmark \citep{cobbe2021gsm8k}. Each problem has a single numeric answer, and the agent receives reward at episode completion.

\subsection{Metrics}

\subsubsection{Reward} For GSM8K dataset, the reward is binary and defined as:

\[
r(y, y^*) = \mathbb{I}[\mathrm{match}(y, y^*)]
\]

where $\mathrm{match}$ is exact equality for GSM8K. The self-review condition is triggered if:

\[
\max_i r_i^{\mathrm{first\ attempt}} < \tau, \; \tau = 1
\]

where the maximum is taken over all $G$ rollouts from the first attempt.

\subsubsection{Learning Efficiency} Beyond the final evaluation reward, we also report learning efficiency as the validation reward trajectory over the training steps. This shows how quickly each method reaches higher reward, which is a key advantage of SRRL.

\subsection{Models and Baselines}

\begin{table}[t]
\centering
\caption{Training hyperparameters for experiments}
\label{tab:training_hyperparameters}
\small
\begin{tabularx}{\columnwidth}{@{}>{\raggedright\arraybackslash}Xcc@{}}
\toprule
\textbf{Hyperparameter} & \textbf{RLVR} & \textbf{SRRL} \\
\midrule
Learning rate & $1 \times 10^{-6}$ & $1 \times 10^{-6}$ \\
KL coefficient ($\beta$) & 0.001 & 0.001 \\
Clipping ratio ($\epsilon$) & 0.28 & 0.28 \\
Rollouts & 6 & 3 per attempt \\
Distillation coeff. ($\lambda$) & --- & 1.0 \\
Optimizer precision & 8-bit AdamW & 8-bit AdamW \\
GPU & H100 & H100 \\
\bottomrule
\end{tabularx}
\end{table}

We train two backbone models using both the RLVR and SRRL frameworks. GRPO is used as an underlying policy gradient optimizer in all cases. First, we used Qwen3-4B, a 4-billion parameter pre-trained language model from the Qwen3 family \citep{yang2025}. The second model used in our experimental study is OLMo-3-7B from the OLMo family \citep{olmo2024furious}.

The RLVR baseline trains the same models with standard reward only GRPO, and there is no self-review, cross-episode memory, or distillation step. In order to ensure that we have a fair comparison of experiments, we equalize the total training compute between the RLVR and SRRL. RLVR uses 6 rollouts per prompt, while the SRRL uses 3 rollouts per attempt across two attempts (i.e. total 6 per episode). Both methods share the same GRPO objective and differ only in the trajectory structure. We apply gradient accumulation with an effective batch size of 32, accumulating gradients across 32 episodes before each optimizer update. This improves gradient stability. For stable training, we employed RL techniques such as clipping, and KL regularization. We train all models using our custom SRRL training loop implemented in Pytorch with the Hugging Face Transformers library. Table~\ref{tab:training_hyperparameters} summarizes all hyperparameters.

\section{Results and Discussion}

\begin{figure*}[t]
\centering
\includegraphics[width=\textwidth]{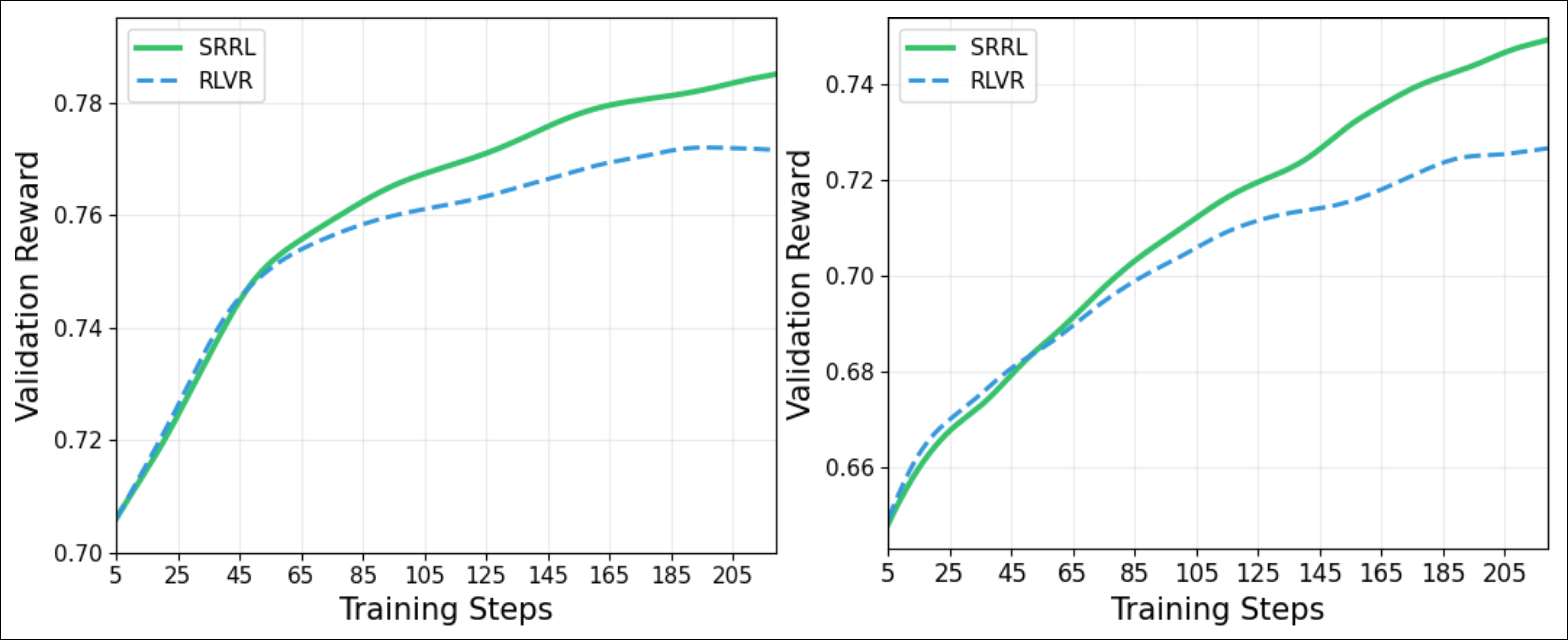}
\caption{Learning efficiency on GSM8K for Qwen3-4B (left) and OLMo-3-7B (right). SRRL achieves higher validation reward than RLVR throughout training and finishes with stronger final performance.}
\label{fig:learning_efficiency}
\end{figure*}

Figure~\ref{fig:learning_efficiency} shows the validation reward trajectories across the 219 training steps for models trained on GSM8K. All the curves are calculated on the fixed held-out validation set of 473 samples, and evaluated every 5 training steps using greedy decoding under the RLVR format constraints.

\subsection{Qwen3-4B}

Both the SRRL and the RLVR frameworks start with a shared rollout reward of approximately 0.71 at the first step. It reflects the model's ability to produce correct answers under the sampled RLVR rollout protocol. It is lower than the 4-shot CoT benchmark, which is reported in the Qwen technical report~\citep{yang2025qwen3}. It is consistent with the known gap between the format-constrained RLVR rollout reward and the standard greedy evaluation. Both curves remain hard to differentiate until approximately step 50. It indicates that both models share similar early gradient signals and the policy has not yet diverged greatly. Afterwards, the SRRL consistently beats RLVR on the 473-sample validation set. After the first 65 training steps, the curve follows a natural deceleration as the model approaches the ceiling of its math-specialized pretraining capacity.

\subsection{OLMo-3-7B}

The OLMo-3-7B curves show a distinct pattern consistent with a general-purpose model on a math-focused benchmark. Both curves begin at a rollout reward of approximately 0.65, which is lower than Qwen3-4B base despite OLMo-3-7B having more parameters. This likely occurs because the OLMo-3-7B pretraining corpus is less math-focused. Early on, both curves remain fused until SRRL begins to diverge. Unlike the Qwen3-4B concave trajectory, OLMo-3-7B exhibits a near-linear rise. This suggests that the model has greater headroom to learn and has not reached a reward ceiling by step 219. SRRL achieves a final validation reward of 0.75, compared with RLVR's 0.72. This corresponds to approximately 10 additional correct answers on the validation dataset.

\subsection{Effects of Self-Review}

The wider SRRL advantage on OLMo-3-7B compared to Qwen3-4B is not coincidental and reveals that self-review benefits models with weaker prior math knowledge. Qwen3-4B already shows strong mathematical priors from pretraining, so the policy quickly learns to generate correct answers via standard reward maximisation. As a result, self-review provides incremental corrections but competes with a capable base policy already. The base policy is more prone to errors in the OLMo-3-7B case. Self-review provides a systematic error-detection signal, and aligns with prior work that shows that iterative self-correction is effective when the base model operates below performance saturation.

The test rewards are closer to the corresponding validation rewards for both models. It's a very consistent train-test gap that shows the absence of validation-set overfitting. SRRL outperforms RLVR on both models, and the OLMo-3-7B achieves a larger absolute improvement than the Qwen3-4B. Notably, Qwen3-4B model achieves higher absolute test performance than OLMo-3-7B despite having fewer parameters. This reversal is the usual parameter-scaling intuition, demonstrated by Qwen3-4B math focus pretraining, which gives it an excellent math reasoning prior. 

\subsection{Implications}

Together, the results support multiple conclusions. First, SRRL surpasses standard RLVR on a general-purpose task and matches specialised base models, which shows promise across two architectures. Also, the magnitude of Self-Review benefit scales inversely with the base model's prior math capability. This finding has practical importance for the model selection. SRRL should be prioritised for deployment on general-purpose base models where the RLVR's reward signal is less informative. Lastly, the qualitatively different learning dynamics between the two models suggest that the shape of reward curves carries significant information on how close the model is to its domain-specific performance ceiling. In future, it may serve as a diagnostic tool for RL training budget allocation.

\begin{table}[t]
\centering
\caption{Final test rewards on the 1,319-sample GSM8K test set.}
\label{tab:gsm8k_test_rewards}
\small
\setlength{\tabcolsep}{8pt}
\renewcommand{\arraystretch}{1.15}
\begin{tabular}{@{}lcc@{}}
\toprule
\textbf{Model} & \textbf{RLVR} & \textbf{SRRL} \\
\midrule
Qwen3-4B      & 0.762 & 0.778 \\
OLMo-3-7B & 0.714 & \textbf{0.742} \\
\bottomrule
\end{tabular}
\end{table}

\section{Conclusion}

We introduced and developed a training framework, known as Self-Review Reinforcement Learning, that embeds an explicit self-review, cross-episode memory and policy distillation steps into the reinforcement learning loop to transform feedback into behavioral improvements. SRRL enables the model to diagnose its failed attempts, learn corrective strategies, and condition a refined second attempt within the same training episode. By combining self-review guidance with selective distillation, SRRL internalizes successful correction strategies into the base policy weights ensuring that improvements persist throughout future training episodes and eliminate the need for self-review at deployment. Across agentic reasoning tasks, the SRRL framework improves the learning efficiency, and produces stronger final polices as compared to standard RLVR baseline.

We evaluated SRRL against RLVR on math reasoning. SRRL directly addressed the latency limitation of inference time reflection methods. Future direction suggests analysing which factor correlates the most with second attempt success the self-review more effective, for example, specificity of error diagnosis, actionability of corrective guidelines, or the current task structure.

\section*{Acknowledgment}
\addcontentsline{toc}{section}{Acknowledgment}

The authors would like to sincerely thank Professor Siva Reddy for his guidance, teaching, and providing a valuable learning experience throughout course.

\bibliography{example_paper}
\bibliographystyle{icml2026}

\newpage
\clearpage

\appendix
\renewcommand{\thesection}{A}
\renewcommand{\thesubsection}{A\arabic{subsection}}

\section{Appendix}
\label{app:appendix_main}

This appendix provides supplementary material for the main paper.

\subsection{Math Prompt}
\label{app:math_prompt}

\begin{table}[H]
\centering
\small
\renewcommand{\arraystretch}{1.15}
\begin{tabular}{p{0.95\linewidth}}
\toprule
\textbf{Template} \\
\midrule
\texttt{Problem: \{question\}} \\[0.3em]
\texttt{You are a math problem solver. Think step by step,} \\
\texttt{then provide your final numeric answer inside} \\
\texttt{<answer>...</answer> tags.} \\
\texttt{Example: <answer>42</answer>} \\
\bottomrule
\end{tabular}
\end{table}

\subsection{Review Prompt}
\label{app:review_prompt}

\begin{table}[H]
\centering
\small
\renewcommand{\arraystretch}{1.15}
\begin{tabular}{p{0.95\linewidth}}
\toprule
\textbf{Template} \\
\midrule
\texttt{You are an expert math tutor reviewing a failed solution.} \\
\texttt{Analyze what went wrong and produce improved solving} \\
\texttt{guidelines inside <prompt>...</prompt> tags.} \\[0.3em]
\texttt{Problem: \{question\}} \\
\texttt{Attempt: \{attempt\}} \\
\texttt{Outcome: INCORRECT (reward = 0.00)} \\
\texttt{[Past successful strategy: \{memory\}]} \\[0.3em]
\texttt{Identify the specific error and produce improved guidelines:} \\
\bottomrule
\end{tabular}
\end{table}

\subsection{GSM8K Dataset Sample}
\label{app:gsm8k_sample}

\noindent\textbf{Problem:} Beth bakes 4 batches of 2 dozen cookies in a week. If these cookies are shared equally among 16 people, how many cookies does each person consume?

\vspace{0.5em}
\noindent\textbf{Solution:} Beth bakes \(4 \times 2\) dozen cookies in total:
\[
4 \times 2 = 8
\]
dozen cookies.

Since each dozen contains 12 cookies, the total number of cookies is:
\[
12 \times 8 = 96
\]

These 96 cookies are shared equally among 16 people:
\[
\frac{96}{16} = 6
\]

\noindent\textbf{Final Answer:} \texttt{<answer>6</answer>}

\subsection{Pseudo code}
\label{app:pseudocode}

\begin{algorithm}[t]
\caption{Self-Review Reinforcement Learning (SRRL) on GSM8K benchmark}
\label{alg:srrl}
\small
\begin{algorithmic}[1]
\STATE \textbf{Inputs:} Language model $\pi_\theta$; dataset of questions $x$; environment returning feedback $f$ and reward $r$; reward threshold $\tau$; KL coefficient $\beta$.
\STATE \textbf{Initialize:} cross-episode memory $m \leftarrow \emptyset$.
\STATE \textbf{repeat}
\STATE \hspace{0.5cm} Sample question $x$ from dataset.
\STATE \hspace{0.5cm} // \textcolor{blue}{First attempt}
\STATE \hspace{0.5cm} Sample first-pass response $y_1 \sim \pi_\theta(\cdot \mid x)$.
\STATE \hspace{0.5cm} Obtain feedback and reward $(f_1, r_1)$.
\STATE \hspace{0.5cm} // \textcolor{blue}{RL update on the first attempt}
\STATE \hspace{0.5cm} Update $\theta$ via $\mathcal{L}_{\text{policy}}(\theta)$ on $y_1$.
\STATE \hspace{0.5cm} // \textcolor{blue}{Gated self-review and second attempt}
\IF{$r_1 < \tau$}
    \STATE \hspace{0.5cm} // \textcolor{blue}{Self-review with cross-episode memory}
    \STATE \hspace{0.5cm} Retrieve relevant memory $m$.
    \STATE \hspace{0.5cm} Sample self-review $\Delta \sim \pi_\theta(\cdot \mid x, y_1, r_1, f_1, m)$.
    \STATE \hspace{0.5cm} Sample second-pass response $y_2 \sim \pi_\theta(\cdot \mid x, \Delta)$.
    \STATE \hspace{0.5cm} Obtain feedback and reward $(f_2, r_2)$.
    \STATE \hspace{0.5cm} Set self-review reward $\tilde{r} \leftarrow r_2$.
    \STATE \hspace{0.5cm} // \textcolor{blue}{Store successful self-review only}
    \IF{$r_2 \ge \tau$}
        \STATE \hspace{1cm} Update memory: $m \leftarrow m \cup \{\Delta\}$.
    \ENDIF
    \STATE \hspace{0.5cm} // \textcolor{blue}{RL update on self-review and second attempt}
    \STATE \hspace{0.5cm} Update $\theta$ via $\mathcal{L}_{\text{policy}}(\theta)$ on $\Delta$ and $y_2$.
    \STATE \hspace{0.5cm} // \textcolor{blue}{Selective policy distillation}
    \IF{$r_2 \ge \tau$}
        \STATE \hspace{1cm} Update $\theta$ via $\mathcal{L}_{\text{distill}}(\theta)$ to train $\pi_\theta$ to produce $y_2$ from $x$ alone.
    \ENDIF
\ENDIF
\STATE \hspace{0.5cm} // \textcolor{blue}{KL regularization}
\STATE \hspace{0.5cm} Apply KL regularization with coefficient $\beta$.
\STATE \textbf{until} converged
\end{algorithmic}
\end{algorithm}

\end{document}